\documentclass[conference]{IEEEtran}

\usepackage[pdftex]{graphicx}
\usepackage{amsmath}
\DeclareMathOperator*{\argmax}{argmax}

\begin{document}

\title{A Sensorimotor Perspective on Grounding\\ the Semantic of Simple Visual Features}

\author{\IEEEauthorblockN{Alban Laflaqui\`ere}
\IEEEauthorblockA{AI Lab, SoftBank Robotics Europe\\
43 Rue du Colonel Pierre Avia, 75015 Paris\\
Email: alaflaquiere@softbankrobotics.com}}

\maketitle

\begin{abstract}
In Machine Learning and Robotics, the semantic content of visual features is usually provided to the system by a human who interprets its content. On the contrary, strictly unsupervised approaches have difficulties relating the statistics of sensory inputs to their semantic content without also relying on prior knowledge introduced in the system. We proposed in this paper to tackle this problem from a sensorimotor perspective. In line with the Sensorimotor Contingencies Theory, we make the fundamental assumption that the semantic content of sensory inputs at least partially stems from the way an agent can actively transform it. We illustrate our approach by formalizing how simple visual features can induce invariants in a naive agent's sensorimotor experience, and evaluate it on a simple simulated visual system. Without any a priori knowledge about the way its sensorimotor information is encoded, we show how an agent can characterize the uniformity and edge-ness of the visual features it interacts with.
\end{abstract}

\IEEEpeerreviewmaketitle

\section{Introduction}
\label{sec:introduction}

Artificial visual perception has made great progress in the last few years, in particular thanks to the development of large images databases and neural network architectures. For three years, computer vision algorithms have even surpassed human performance in classification tasks on specific databases \cite{he2016deep} 
Despite these impressive achievements, current artificial vision systems still exhibit important limitations. As exemplified by the existence of adversarial examples, their high-performances can prove surprisingly brittle \cite{goodfellow2014explaining}. But, even more concerning for the Developmental Robotics community, another strong limitation of these systems is their lack of autonomy. Indeed, current efficient machine learning systems are supervised. They rely on humans to collect and pre-process the adequate task-related data, to define a suitable network architecture, and most importantly to provide an interpretation of the semantic content of the visual scene in the form of labels or rewards.

The question of how to build a completely autonomous artificial vision system remains open. In particular, how can a robot create or discover the semantic content of the visual input it receives. Unsupervised approaches of the problem have been proposed \cite{lesort2018state}, but they still eventually require a human to interpret the patterns that have been statistically extracted from the data.
The grounding of semantics is a deep philosophical question that has arguably been investigated for centuries, and that roboticists have practically bumped into since the early years of artificial intelligence. Quite evidently, it will not be solved easily. We can nonetheless start addressing the problem by looking at simple problems that could provide some insight on how to solve more complex ones, and in particular on how humans perceive their environment autonomously.

\begin{figure}[!t]
\centering
\includegraphics[width=1\linewidth]{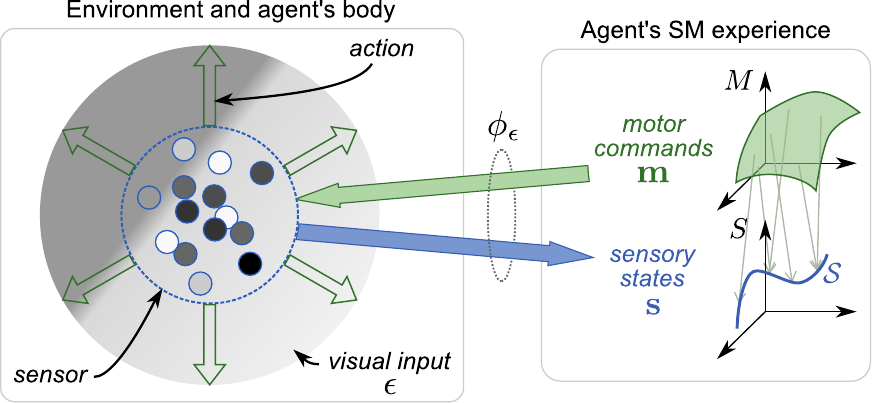}
\caption{Experimental setup: A naive agent explores visual inputs and characterizes them through the sensorimotor invariants they induce.}
\label{fig:principle}
\end{figure}

We follow such an approach and investigate the grounding of the perception of simple visual features.
We base our study on the Sensori-Motor Contingencies Theory (SMCT), a theory of perception that was introduced with a particular consideration for visual experience \cite{o2001sensorimotor}. This theory suggests that the subjective experience of perception emerges from regularities in our sensorimotor flow. More precisely, it argues that perception does not come directly from the processing of passive sensory inputs, but from the knowledge of the way one's actions would transform these sensory inputs.
This philosophical perspective has multiple interesting consequences for robotics, and in particular autonomous and developmental systems. It suggests that a robot can acquire perceptive abilities by actively exploring its environment and identifying regularities in its sensorimotor experience. But more interestingly, it suggests that the subjective perceptive experiences themselves can be characterized by the properties of the sensorimotor regularities they are associated with.
A typical example of this idea is the one of a line, or more generally an edge. There is a sensorimotor regularity when one looks at an edge: regardless of the way the sensorimotor information is encoded, actions that move the eye generate sensory variations, except when the eye moves along the edge. This specific sensorimotor invariant characterizes the visual input through the way one can interact with it, and independently from the static properties of the visual input itself.
\\
In this paper, we propose to investigate the practical relevance of this philosophical claim by evaluating the sensorimotor invariance associated with simple visual features. To do so, we propose a mathematical formalization of the problem, as well as simple simulations of an agent exploring its environment with a small retina-like sensor (See Fig.~\ref{fig:principle}).

Previous works have developed approaches inspired by the SMCT. They studied different components of perceptive experience such as space \cite{terekhov2013space}, color \cite{witzel2015determines}, objects \cite{maye2011discrete}, field of view \cite{laflaquiere2017grounding}, tactile space \cite{marcel2017building}, or auditory space \cite{bernard2012sensorimotor}, or containment \cite{hay2018behavior}. Despite some of them being in part related to visual experience, none directly addresses the problem of characterizing visual features. Nonetheless, a similar approach has previously been proposed in \cite{choe2010motor}. Our work differs from it in that we propose a mathematical formalization of the visual sensorimotor invariances, instead of casting the problem in a Reinforcement Learning framework which relies on a hand-designed reward function.

In the following sections, we introduce a mathematical formalization of the problem, propose a method to identify sensorimotor invariants induced by simple visual features, and evaluate it on simple experiments. Finally we discuss our results and the practicality and limitations of the approach.

\section{Problem}
\label{sec:problem}

In our study, we avoid as much as possible any bias that is usually introduced in the processing of sensorimotor experiences.
To do so, we consider naive ({\it tabula rasa}) agents which do not have any a priori knowledge about their environment, nor the sensorimotor apparatus they use to explore it. The \emph{agent} itself is considered to be the information processing system which only access the environment indirectly through the interface formed by its physical sensors and motors.
(see Fig.~\ref{fig:principle}).
As a consequence, the agent has to estimate the environment's properties by looking at the instantaneous sensory state and motor states that it receives and generates. We respectively define them as:
$$\mathbf{s} = [s_1, s_2,\dots, s_{N_s}]^T \text{ and } \mathbf{m} = [m_1, m_2,\dots, m_{N_m}]^T,$$
where $s_i, i\in\{1,\dots,N_s\}$ is the individual sensation produced by the $i$-th sensor, and $m_j, j\in\{1,\dots,N_m\}$ is the individual motor command sent to the $j$-th motor. We denote $S$ and $M$ the vector spaces in which $\mathbf{s}$ and $\mathbf{m}$ live.\\
Although this formalization is relatively  general, we limit in this paper our study to visual sensations and assume that the agent is equipped with a type of visual sensor. This way, each sensation $s_i$ can be thought of as produced by an individual cell (cone or rod) in a retina or a camera (pixel). The topological organization of those elementary sensors, as well as the way they encode the information are however unspecified.
Similarly, we assume that the motor commands correspond to displacements of the sensor in the visual scene, locally akin to translations in the plane.
Based on this basic formalism, we address the question of the identification of properties which could characterize sensory inputs.

\subsection{Passive approach}
\label{sec:passive approach}

In the absence of prior knowledge, or external inputs (label, reward), the common way to address the problem is to perform a statistical analysis of a collection of static sensory inputs $\{ \mathbf{s}_k \}$ 
. This way, one can for instance estimate the probability of occurrence of a sensory input. One can also evaluate the correlation between the different components $s_i$ of the sensory state $\mathbf{s}$. This type of approach leads to the extraction of sensory statistics which can be very useful for bootstrapping the solving of computer vision tasks \cite{erhan2010does}. For instance, analysis images from the internet, it
can create features, or representations, specific to "cats" and "faces", as presented in \cite{le2013building}. Yet, a human is still required to interpret the semantic content of these statistical representations. For instance, such a system can capture the fact that an oriented edge or a uniform input are highly probable, but cannot make explicit in what way each of them is particular.

In order to estimate semantic content from static sensory inputs, one needs to incorporate prior knowledge into the system. For instance, it is possible to evaluate the 'uniformity' of a visual input \emph{if} the excitation function $s_i = f_i(e_i)$ of each sensor is provided. This way, one can trace back the local state of the environment $e_i$ captured by the $i$-th sensor from the sensation $s_i$, and compare the states $e_i$ to see how much they differ from one another.
Another example is the possibility to estimate the presence of an edge \emph{if} the excitation functions $f_i$ are known, and \emph{if} the topological organization of the sensors is known. In such a case, one can evaluate if two \emph{linearly separable} sets of pixels encode two different environmental states $e_i$.
Besides the prior knowledge about the sensory apparatus that these evaluations require, it is important to notice that they are defined by a human who deems them meaningful. The sensory inputs in themselves do not exhibit particular interesting properties for the agent. For instance, given different unspecified excitation functions $f_i$, a uniform visual input would be encoded as a vector of different sensations $s_i$ which has no more semantic content for the agent than a random visual input would have for a human.

\subsection{Active approach}
\label{sec:active approach}

As suggested by the SMCT, a sensorimotor approach is possible to characterize visual inputs. Unlike the typical computer vision approach which relies on a collection of static sensory inputs, it takes into account the spatio-temporality of the data and the link between the motor and sensory streams.
Adding this motor component to the problem expends the space (now sensorimotor) in which the data can be analyzed. In particular, it is possible to look at how actions transform sensory inputs. For an autonomous agent that needs to \emph{act} in the world, a strong argument can be made that regularities in the way actions can transform sensations are more useful to extract than passive regularities in the sensory states only.

We denote $\phi$ the unspecified sensorimotor function which maps the motor commands to the sensory states:
\begin{equation}
\mathbf{s} = \phi_{\epsilon}(\mathbf{m}).
\end{equation}
It is parametrized by $\epsilon$ which represents the state of the environment the agent is currently interacting with, that we also refer to as \emph{visual input} in the context of this paper. This visual input is to be distinguished from the sensory input $\mathbf{s}$ that $\phi_{\epsilon}$ generates.
The mapping $\phi_{\epsilon}$ is characteristic of the visual input $\epsilon$, as the agent's sensorimotor experience varies depending on $\epsilon$. In particular, some mappings $\phi_{\epsilon}$ can induce invariants in the sensorimotor experience.

We hypothesize that simple visual features can be characterized through their associated sensorimotor invariants.
As mentioned in Section~\ref{sec:introduction}, an edge is for example associated with a specific sensorimotor invariance: regardless of the way sensory and motor information are encoded, there is a set of motor commands which leave this sensory input unchanged. Moreover, the orientation of the edge is characterized by the motor commands to which the sensory input is invariant. Similarly, a uniform visual input exhibits this kind of invariance for any motor command.

Formally, the function $\phi_{\epsilon}$ exhibits such a pointwise invariant if there exists an $\mathbf{m}$ such that:
\begin{equation}
s = \phi_{\epsilon}(0) = \phi_{\epsilon}(\mathbf{m}),
\label{eq:invariant}
\end{equation}
where $m=0$ is arbitrarily considered as a reference motor state\footnote{Given the type of visual interaction we consider, we can assume that any visual input $\epsilon$ can be experienced when $\mathbf{m}=0$.}.
Despite not having direct access to $\phi_{\epsilon}$, the agent can discover this invariant by analyzing its sensorimotor experience.
In contrast with supervised approaches in which the learning is guided by human inputs, identifying such an invariant is intrinsically interesting, as it characterizes the kind of interaction the agent can have with the environment. Moreover, it suggests a kind of abstraction from the sensory states themselves, as different states (for examples, edges with the same orientation but different colors) can share the same invariant.

\subsection{Experimental setup}
\label{sec:experimental setup}

In the following sections, we investigate how a naive agent can extract sensorimotor invariants of the type of Eq.~\ref{eq:invariant} and characterize simple visual features this way.
To do so, we propose a simple experiment simulating the exploration of visual inputs by an agent (see Fig.~\ref{fig:principle}).

The environment explored by the agent consists of visual inputs $\epsilon$. They can be conceptualized as functions $v_{\epsilon}$ which take as input a position $(x,y)$ in the plane, and generates an output denoted $e$:
\begin{equation}
e = v_{\epsilon}(x,y).
\end{equation}
We arbitrarily define the functions $v_{\epsilon}$ such that their potential output space is limited to $[0,1]$.
Typically, a grayscale image captured by a camera would be a sampling of such a function over a regular grid, with each pixel taking values in $[0,1]$.
The size of each visual input is set to $10\times 10$ units\footnote{Any unit of length could be considered, as all distances in the simulation are relative.}, as the function is defined over the subspace $x,y \in [-5,5]$.

The agent is equipped with a generic visual sensor made up of $N_s = 25$ elementary cells spread in a disk of diameter $4$ units. For each cell, the direction and distance to the center of the disk are randomly drawn from uniform distributions at the beginning of the simulation, yielding a non-homogeneous topological organization of the overall visual sensor.
Each cell $i$ captures the local visual input
\begin{equation}
e_i = v_{\epsilon}(x_i, y_i)
\end{equation}
at the position $(x_i, y_i)$ of the cell in the plane. The excitation function $f_i$ of each cell is independently defined as an arbitrary continuous function:
\begin{equation}
s_i = f_i(e_i) = \alpha_i e_i^2 + \beta_i e_i + \gamma_i,
\end{equation}
with $\alpha_i, \beta_i, \gamma_i$ as fixed parameters drawn at the beginning of the simulation. It prevents a direct comparison of the different sensations $s_i$.

The agent is equipped with $N_m=2$ motors which respectively control the horizontal and vertical displacements $(\Delta x, \Delta y)$ of the sensor in the plane. The sensor can be moved continuously in the environment, which effectively changes the position of each cell $i$ to $(x_i+\Delta x, y_i + \Delta y)$.
The motor exploration of each visual input is limited to a disk of $6$ units diameter. For each displacement of the sensor, the direction and amplitude are randomly drawn from uniform distributions. This choice is arbitrary and could be replaced by any other sampling of the motor space.\\
During the motor exploration, we assume that the agent is re-centered on the visual input after each movement. This constraint can seem artificial, but it ensures that only a single environmental state $\epsilon$ is locally explored and characterized. It is to be seen as a simplification of a more realistic scenario in which the agent iteratively moves around and encounters a different environmental state $\epsilon$ at each iteration. The constrained motor exploration we propose here is simply the collection of all the sensorimotor experiences that such a free agent would get when encountering a specific visual input $\epsilon$.

\begin{figure*}[!t]
\centering
\includegraphics[width=\linewidth]{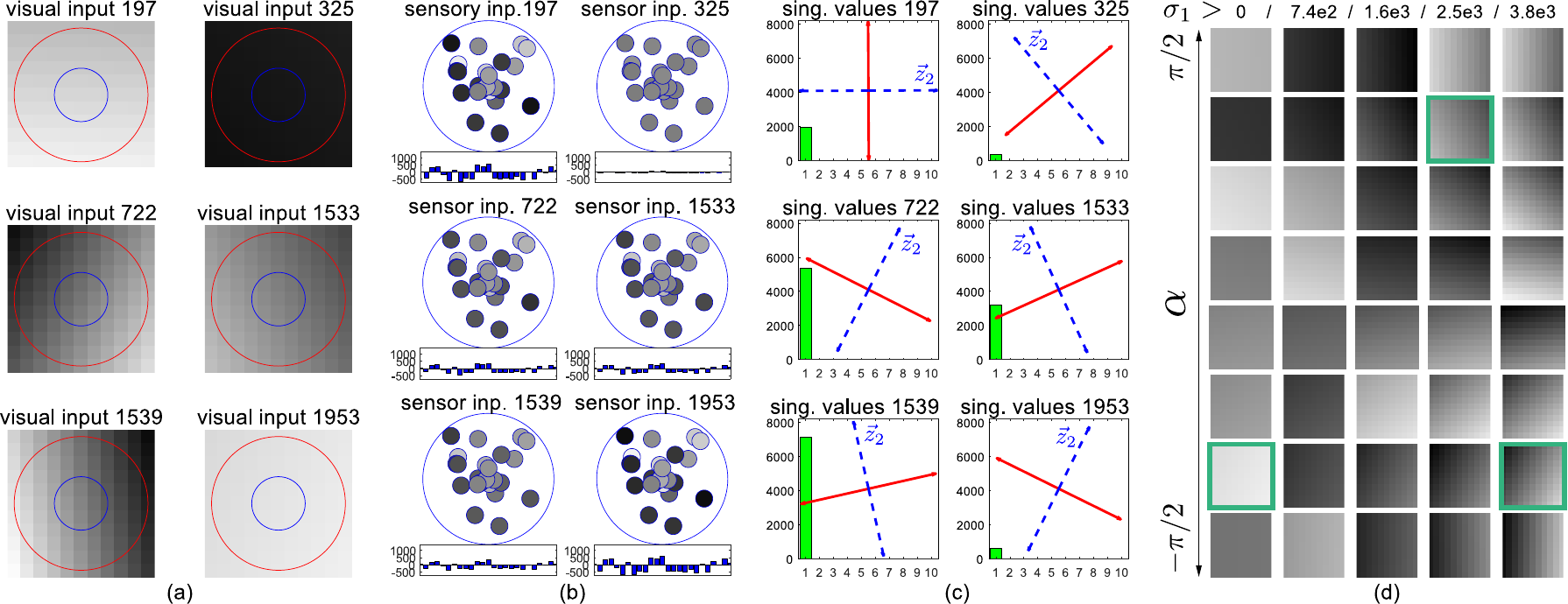}
\caption{Experiment 1: (a) Examples of visual inputs represented as regular grids of pixels. The sensor limits is shown in blue and the limit of the motor exploration in red. (b) The corresponding sensory encoding, represented as cells and sensory vector accessible to the agent. (c) Singular values of $D_s$, as well as the first (red) and second (dashed blue) motor directions defined by $R_{:1} and R_{:2}$. (d) Sub-sampling of the $2000$ visual inputs organized according to $\sigma_1$ and $\vec{z}_2$. Three visual inputs present in panel (a) are outlined in green.}
\label{fig:exp1}
\end{figure*}

\section{Linear sensorimotor function}
\label{sec:linear sensorimotor function}

In any non-trivial system, the function $\phi_{\epsilon}$ is very complex. It implicitly embodies all the unknown properties of the visual input and the agent's sensorimotor apparatus. We can however study it locally to extract its potential invariants.

\subsection{Linear approximation}
\label{sec:linear approximation}

Let's assume that the function $\phi_{\epsilon}$ is smooth (in the mathematical analysis sense), and re-express it locally as a linear function:
\begin{equation}
\mathbf{s} = A_{\epsilon}(\mathbf{m}) + B_{\epsilon}
\end{equation}
where
$A_{\epsilon}$ is a $N_s \times N_m$ matrix, and $B_{\epsilon}$ is a bias vector of size $N_s$.
Assuming that the number of sensors $N_s$ is at least equal to the number of motors $N_m$, and that the motors are independent, the rank of $A_{\epsilon}$ is at most equal to $N_m$. It can however be smaller if there exists a direction in $M$ along which $A_{\epsilon}$ does not induce any sensory change in $S$.

In practice, the agent does not have access to $A_{\epsilon}$, but only to $\mathbf{m}$ and $\mathbf{s}$. Nonetheless, if we denote $\mathcal{S} \subset S$ the image subspace of $A_{\epsilon}$, the intrinsic dimension of $\mathcal{S}$ is equal to the rank of $A_{\epsilon}$. It is thus possible to randomly generate a $N_m \times K$ matrix $D_m$ of $K$ samples $\mathbf{m}_k$ in $M$ to create a sampling matrix $D_s$ of size $N_s \times K$ containing the resulting sensory variations:
$$\Delta \mathbf{s}_k = \phi_{\epsilon}(\mathbf{m}_k) - \phi_{\epsilon}(0),$$
and to perform a singular value decomposition (SVD) of $D_s$:
\begin{equation}
D_s = L \Sigma R^T,
\end{equation}
where $L$ is an $N_s \times N_s$ unitary matrix, $R$ is an $K \times K$ unitary matrix, and $\Sigma$ is a $N_s \times K$ diagonal matrix with the singular values $\sigma_i$ of $D_s$ in decreasing order.
The number of significant (non-null) singular values correspond to the intrinsic dimension of $\mathcal{S}$, and thus to the rank of $A_{\epsilon}$.
Moreover, the first columns of $R$ (right-singular vectors) associated with those significant singular values correspond to the combinations of motor samples in $D_m$ which induce sensory changes, while other columns of R correspond to the combinations of samples in $D_m$ which leave the sensory state invariant. In other words, based on the sampling $\{D_m, D_s\}$, we can estimate the rank of $A_{\epsilon}$ associated with a visual input $\epsilon$, as well as the potential motor commands $\mathbf{m}$ which leave the sensory state $\mathbf{s}$ invariant. Those properties can be used to intrinsically characterize the visual input $\epsilon$.

\begin{figure*}[!t]
\centering
\includegraphics[width=1\linewidth]{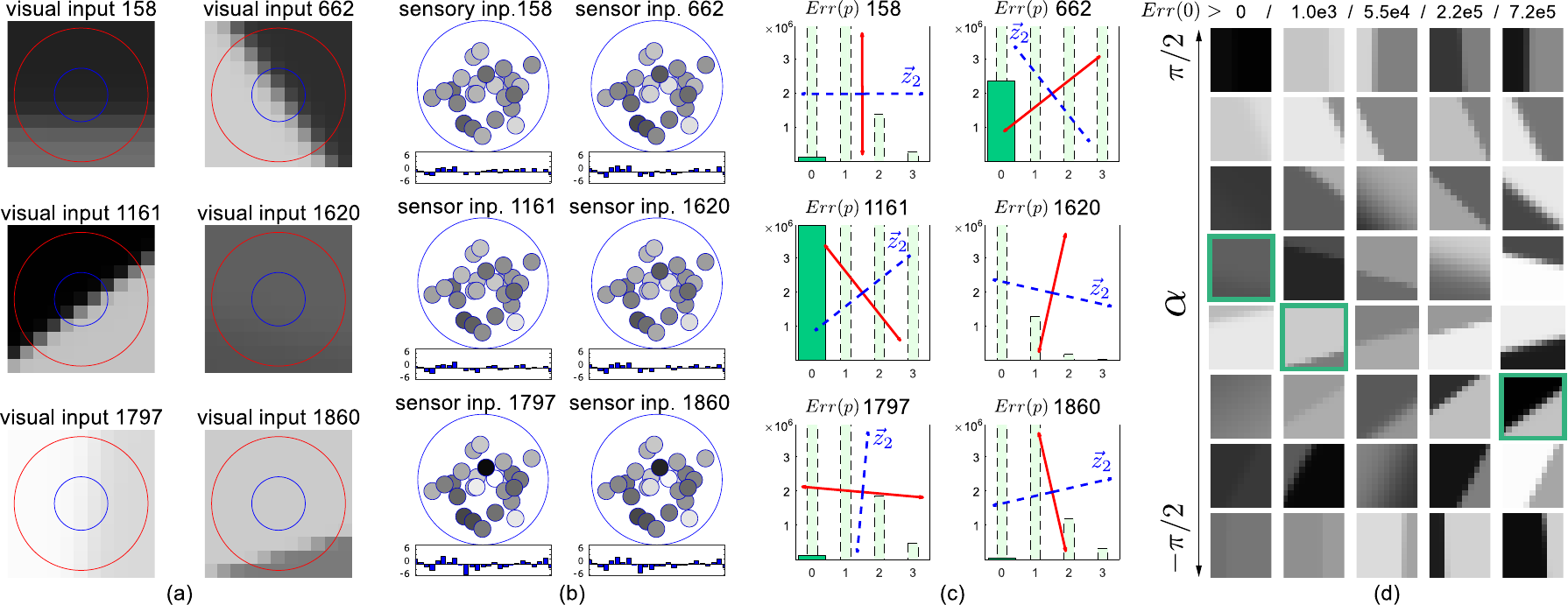}
\caption{Experiment 2: (a) Examples of visual inputs, as in Fig.~\ref{fig:exp1}. (b) The corresponding sensory encoding. (c) Projection errors $Err(p)$ of $D_s$, and the first (red) and second (dashed blue) motor directions defined by $R_{:1} and R_{:2}$. Singular values of $D_s$ are displayed with dashed bars (scaled up). (d) Sub-sampling of the $2000$ visual inputs organized according to $Err(0)$ and $\vec{z}_2$. Three visual inputs present in panel (a) are outlined in green.}
\label{fig:exp2}
\end{figure*}

\subsection{Experiment 1}
\label{sec:experiment 1}

In the first experiment, we simulate linear sensorimotor mappings $\phi_{\epsilon}$. To do so, we create visual $2000$ features such that $v_{\epsilon}$ is linear.
As illustrated in Fig~\ref{fig:exp1}a, they correspond to gradients with various orientations, slopes, and biases. We also ensure that all excitation functions $f_i$ are linear by drawing $\beta_i$ and $\gamma_i$ from a uniform distribution $\mathcal{U}(-1e3,1e3)$, but setting $\alpha_i$ to $0$.
Examples of the resulting sensory encoding of visual features are illustrated in Fig.~\ref{fig:exp1}b.

The agent explore each visual input with $K=1000$ random motor commands. The sensory sampling $D_s$ generated this way is analyzed through a SVD. Potential sensorimotor invariances are estimated by looking at the number of significant singular values $\sigma_i$: 
\begin{itemize}
\item none means the sensory input is invariant to any motor command,
\item one means the sensory input is invariant to one direction in the motor space,
\item two means the sensory input is not invariant to any motor command.
\end{itemize}
Note that because the motor space is $2$D, no more than two $\sigma_i$ can be significant.
For a given non-significant $\sigma_i$, the motor direction associated with the related invariance is:
\begin{equation}
\vec{z}_i = \frac{D_m  R_{:i}}{||D_m  R_{:i}||}.
\end{equation}

Results of the simulation are presented in Fig.~\ref{fig:exp1}c.
For all visual inputs, only one singular value is significant. This result is expected as the visual gradients all exhibit one invariant. Moreover, $\sigma_1$ tends towards $0$ for visual inputs which are close to uniform.
We can also see that the direction of the invariance is correctly estimated via the SVD.\\
Despite having no information about the encoding of its sensorimotor information, the agent is thus able to characterize the 'uniformity' visual inputs via the value of $\sigma_1$, and their 'edge-ness' via the value of $\sigma_2$ and its corresponding right-singular vector $R_{:2}$.
Note that, due to the linearity of the visual input, there cannot exist two simultaneous dimensions of variations for $v_{\epsilon}$. As a consequence, the second singular value $\sigma_2$ is always non-significant.\\
The last panel of Fig.~\ref{fig:exp1} displays a larger set of visual patches characterized by the agent. They are organized horizontally accordingly to their estimated uniformity, and vertically according to their estimated direction of motor invariance. We can see that the agent can build this way a topological representation of the visual inputs which reflect their invariances.

\section{Extension to non-linear functions}
\label{sec:extension to non-linear functions}

Although mathematically convenient to manipulate, the linear approximation of $\phi_{\epsilon}$ by $A_{\epsilon}$ rarely stands in realistic scenarios. For real visual interactions, the sensorimotor mapping $\phi_{\epsilon}$ can be strongly non-linear.
The linear method proposed to characterize sensory inputs can however be extended to a non-linear setting by taking inspiration from differential geometry.

\subsection{Analyzing the sensory manifold}
\label{sec:Analyzing the sensory manifold}

The subspace $\mathcal{S}$ spanned by a smooth non-linear function $\phi_{\epsilon}$ is a manifold whose intrinsic dimension can be thought of as a non-linear extension of the rank of $A_{\epsilon}$.
Instead of estimating the rank of the sample matrix $D_s$ produced by a non-linear function, one can thus estimate its intrinsic dimension. Numerous methods could be considered to perform such an estimation. We propose in this work to use the Curvilinear Component Analysis (CCA) \cite{XXX} to project the data $D_s$ in spaces of lower dimensions $p < N_s$, and to monitor the projection error to determine the smallest dimension $p^*$ for which the error is non-significant.
Let $Y_s^{(p)}$ denote the projection of $D_s$ in dimension $p$, and $Err(p)$ be the projection error in dimension $p$. The intrinsic dimension of $D_s$ is estimated as:
\begin{equation}
p^* = \argmax_{p} \left( \frac{Err(p-1)}{Err(p)} \right).
\label{eq:dim}
\end{equation}
Note that it is impossible to estimate $p^*=0$ with this method.

To determine the motor commands to which the non-linear function $\phi_{\epsilon}$ might be invariant, one has to estimate its zero set. This is a more complicated problem as there is no common way to determine the zero set of unspecified non-linear functions. In this work we propose to use $Y_s^{(p^*)}$, the optimal low-dimensional projection of $D_s$, as a linear approximation of the unfolded manifold.
Like in the linear case, we can then perform a SVD of $Y_s^{(p^*)}$: 
\begin{equation}
Y_s^{(p^*)} = L \Sigma R^T,
\end{equation}
and determine the motor combinations that do not generate sensory changes by looking at the motor combinations defined by the right-singular vectors in $R$ associated with non-significant singular values.

\subsection{Experiment 2}
\label{sec:experiment 2}

In the second experiment, we simulate non-linear sensorimotor mappings $\phi_{\epsilon}$. To do so, we create $2000$ non-linear visual feature by generating simple gradients and passing them through $\tanh$ functions with random slopes and biases.
As illustrated in Fig.~\ref{fig:exp2}a, they correspond to sharper edges for which the visual input is not linear with regards to the position $(x,y)$.
Moreover, non linear excitation functions $f_i$ are generated by independently drawing their parameters $\alpha_i, \beta_i, \gamma_i$ from a normal distribution $\mathcal{N}(0,1)$. Examples of the resulting sensory encoding of visual features are illustrated in Fig.~\ref{fig:exp2}b.

As in the previous experiment, the agent explore each visual input with $K=1000$ motor commands. The resulting sampling $D_s$ is then projected in low dimension via a CCA with $p \in \{0,1,2,3\}$.
Potential sensorimotor invariances are identified by looking at the estimated intrinsic dimension $p^*$ of the sensory manifold, and the right-singular vector $R_{:(p^*+1)}$ associated with the first non-significant dimension $p^*+1$.

Results of the simulation are presented in Fig.~\ref{fig:exp2}c.
For all visual inputs in the simulation, the intrinsic dimension $p^*$ is estimated equal to $1$ by the agent, whereas $D_s$ exhibits a greater number of significant $\sigma_i$. The non-linear analysis of the manifold's dimensionality is thus conclusive as all visual features exhibit at least one invariant.
Note that the intrinsic dimension of $D_s$ might be equal to $0$ for some uniform visual inputs, but our dimension estimation method is unable to detect it (see Eq.~\eqref{eq:dim}). Uniformity can however be estimated by looking at $Err(0)$, as displayed in Fig.~\ref{fig:exp2}d.
Moreover, we can see that the direction of the invariance in the motor space is correctly estimated via the SVD of $Y^{1}_s$.
The naive agent is thus able to characterize the uniformity of the visual features, as well as their edge-ness and orientation, even when the sensorimotor mapping $\phi_{\epsilon}$ is non-linear.
Figure~\ref{fig:exp2}d displays a larger set of visual patches organized according to their uniformity and orientation. We can see that the agent can build a topological representation of the visual inputs characterizing their invariances.

\section{Discussion}
\label{sec:discussion}

We presented in this work a mathematical formalization and a preliminary experimental evaluation of the sensorimotor characterization of simple visual features. In line with the SMCT, we propose that visual inputs can be characterized, without any a priori knowledge, by looking at the properties of the sensorimotor regularities they induce.
With the formalization and simple simulation proposed in this paper, we have  shown how a naive agent can characterize their uniformity and edge-ness by locally exploring visual inputs and detect their potential sensorimotor invariants. Based on those invariants, the agent can internally build its own low-dimensional topological representation of the visual inputs it encounters in the environment.
In contrast with typical passive analysis of sensory inputs, this representation intrinsically informs the agent on its ability to transform (or not) the related sensory input; a knowledge that would be directly useful for planning future actions.
Such a sensorimotor characterization of visual inputs also seems to lead to basic abstraction. For instance, the uniformity of visual inputs can be characterized independently from their intensity (light or dark). Similarly, edges between areas of different intensity can be clustered in more abstract groups based on their orientation.

Despite these encouraging preliminary results, many challenges need to be overcome before a complete formalization of the grounding of visual experience is proposed.
Firstly, the system simulated in this work is simplistic. It only represents a local interaction between a simple visual feature and what would correspond to a small receptive field in our large field of view. Understanding how this sensorimotor approach can be scaled up to multiple receptive fields observing a visual scene in parallel is a natural question to investigate in the future. Some preliminary work has already been proposed in this direction \cite{laflaquiere2017grounding}. This will also naturally raise the question of motor actions of greater amplitude, and how information can circulate between receptive fields to deal with greater displacements.
Secondly, sensory ambiguity has not been addressed in this work. Indeed, it is possible that multiple environmental states $\epsilon$ generate the same sensory experience $\mathbf{s}$ for a given motor state. This means that the agent's sensory experience is potentially ambiguous, and that a probabilistic extension of the current formalism is necessary. Given a sensory input in its receptive field, the agent would thus estimate a distribution over the probable associated environmental states $\epsilon$, that it could disambiguate by performing a motor action or collecting information from surrounding receptive fields.
Thirdly, we assumed the existence of perfect pointwise invariants in our formalism. However, real sensorimotor interactions can be noisy and break this assumption. This problem could be tackled by instead looking for setwise invariants, for which the set corresponds to a small neighborhood around the sensory state  $\mathbf{s}$. This way, invariants could be identified by looking for motor actions which map a set of noisy data to itself.
Finally, a long term goal is to investigate how the unsupervised capturing of sensorimotor invariants can be coupled with a planning or reinforcement learning module in order to perform guided interactions with the world. In particular, one would need to formalize how the simple abstraction induced by our approach is beneficial for guiding an agent towards its goal.

\bibliographystyle{IEEEtran}
\bibliography{biblioICDLSemantic}

\end{document}